%% file: main.tex
\documentclass{article}

\PassOptionsToPackage{round,authoryear}{natbib}
\usepackage[preprint]{neurips_2026}
\input{preamble}

\title{Property-Based Testing for Robust Softmax and Cross-Entropy}
\author{Muhammad Maaz}
\date{}

\begin{document}

\maketitle

\begin{abstract}
  A recent paper of \citet{barron2026fix} identified two numerical issues with the softmax and cross-entropy functions in virtually all standard machine learning libraries (e.g., PyTorch and JAX). We show that these issues can be recovered through simple property-based tests in Hypothesis. However, our tests identify an additional edge case where their implementation fails. We then propose an even more robust softmax implementation which passes these tests and also supports use in causal attention masking. We compare our implementation with theirs and discuss some of the differences. Overall, this work shows that property-based testing can be a powerful tool for discovering and fixing numerical issues in machine learning libraries. Our implementations and property-based tests can be found at \url{https://github.com/mmaaz-git/softmax-fix}.
\end{abstract}

\section{Introduction}
\label{sec:introduction}

Suppose we have logits $\vect{z} \in \R^K$. It is common to compute $\softmax(\vect{z}) = e^{\vect{z}} / \sum_{i=1}^K e^{z_i}$ by subtracting the maximum logit \citep{higham2002accuracy}, which is more numerically stable. More precisely, we compute the log-softmax as follows:

\begin{align*}
  \logsoftmax(\vect{z}) = (\vect{z} - \max_i z_i) - \log \sum_{i=1}^K e^{z_i - \max_i z_i}
\end{align*}

Then, we can compute the softmax as $\softmax(\vect{z}) = e^{\logsoftmax(\vect{z})}$, or directly by applying the max-shift trick:

\begin{align*}
  \softmax(\vect{z}) = e^{\vect{z} - \max_i z_i} / \sum_{i=1}^K e^{z_i - \max_i z_i}
\end{align*}

The max-shift trick is widely used in practice, e.g., in PyTorch and JAX.

Given logits $\vect{z}$ and a target probability vector $\vect{p} \in \Delta^{K-1}$, the soft-target cross-entropy is $\CE(\vect{z}, \vect{p}) = - \sum_{i=1}^K p_i [\logsoftmax(\vect{z})]_i$. When $\vect{p} = \vect{e}_t$ is a one-hot vector corresponding to target class $t$, this simplifies to the hard-target cross-entropy: $\CE(\vect{z}, \vect{e}_t) = -[\logsoftmax(\vect{z})]_t$. 

Recently, \citet{barron2026fix} observed that even with the max-shift trick, two possible numerical issues can arise: (1) softmax collapse, where, as the model becomes more confident in the correct class, the gradient of the hard-target cross-entropy with respect to the target logit incorrectly rounds to 0, and (2) softmax cancellation, where the $L^2$ norm of the gradient of the soft-target cross-entropy incorrectly plateaus at a non-zero value as the student logits approach the teacher logits. They claim that these issues were found and fixed using Gemini Flash 3.7 \citep{barron2026softmax_tweet}. The problem of softmax collapse in grokking was previously studied by \citet{prieto2025grokking}.

However, we show that these issues can also be found via \textit{property-based testing}. Property-based testing is a type of software testing where an abstract property of the system under test is specified, and then random inputs are generated and checked against this property. Properties can pertain to single functions (e.g., that for all inputs, the output should be non-negative), or, more powerfully, to multiple functions (e.g., that encoding and decoding yields the original input). Property-based testing began with QuickCheck in Haskell \citep{claessen2000quickcheck}, and has since been adopted by many other programming languages, including in Python, in the Hypothesis library \citep{maciver2019hypothesis}. Both of these issues expose failures of key mathematical properties of the softmax and cross-entropy functions, which take no more than a few lines of code to test. Property-based testing is increasingly being adopted in the test suites of popular software libraries, like NumPy, and in fact has been used to discover bugs in NumPy and other libraries \citep{maaz2025agentic}.

However, we will show that \citet{barron2026fix}'s fix still fails. We propose a better fix which is even more robust, and show that this allows the new softmax to be used in the setting of causal attention masking. We release our implementations and tests at the codebase at \url{https://github.com/mmaaz-git/softmax-fix}.

\section{Rediscovering the Bugs with Property-Based Testing}
\label{sec:rediscovering}

For a logit vector $\vect{z}$, let $\vect{q} = \softmax(\vect{z})$ and denote its Jacobian by $J(\vect{z})$, with entries $J_{ij} = \partial q_i / \partial z_j$. Fix an output index $i \in \{1,\ldots,K\}$. It is well-known that:

\begin{align*}
  \frac{\partial q_i}{\partial z_i} &= q_i (1 - q_i), \\
  \frac{\partial q_i}{\partial z_j} &= -q_i q_j, \quad j \neq i.
\end{align*}

Thus, the entries of the gradient of each output probability sum to zero:

\begin{align*}
  \sum_{j=1}^K \frac{\partial q_i}{\partial z_j} = q_i (1 - q_i) - \sum_{j \neq i} q_i q_j = q_i (1 - q_i) - q_i (1 - q_i) = 0.
\end{align*}

Hence, every row of the Jacobian $J(\vect{z})$ sums to zero in exact arithmetic. Failure of this identity is related to the softmax collapse phenomenon as described by \citet{barron2026fix}.

To see why, suppose the target logit $z_t$ dominates, and define the margin $m$ by
\begin{align*}
  m = z_t - \log \sum_{j \neq t} e^{z_j},
  \qquad q_t = \frac{1}{1 + e^{-m}}.
\end{align*}
For sufficiently large $m$, $1 + e^{-m}$ rounds to $1$. Using hats to denote computed quantities, we then have $\widehat q_t = 1$ and $\sum_{j \neq t}\widehat q_j > 0$. The standard softmax backward formula therefore gives
\begin{align*}
  \widehat J_{tt} &= \widehat q_t(1-\widehat q_t) = 0, \\
  \widehat J_{tj} &= -\widehat q_t\widehat q_j = -\widehat q_j,
    \quad j \neq t, \\
  \sum_{j=1}^K \widehat J_{tj} &= -\sum_{j \neq t}\widehat q_j < 0.
\end{align*}

This also explains the corresponding hard-target cross-entropy failure. For $L = \CE(\vect{z},\vect{e}_t) = -\log q_t$, the chain rule gives
\begin{align*}
  g_j = \frac{\partial L}{\partial z_j}
      = -\frac{1}{q_t}J_{tj}.
\end{align*}
Propagating the erroneous Jacobian row through this expression yields $\widehat g_t = 0$ and $\widehat g_j = \widehat q_j$ for $j \neq t$, so $\sum_j\widehat g_j > 0$. The standard cross-entropy backward formula $\widehat g_t = \widehat q_t - 1$ exhibits the same failure. In exact arithmetic, however, the target gradient is
\begin{align*}
  g_t = -\frac{e^{-m}}{1+e^{-m}} = -\sum_{j \neq t}q_j,
\end{align*}
which balances the non-target gradients.

In fact, it is possible to compute when this softmax collapse occurs, based on the precision of the numerical representation. Recall that the collapse arises due to floating point numbers near $1$. Recall that $q_t = 1 / (1 + e^{-m})$. If the format has $b$ significant bits, then the next representable number above $1$ is $1+2^{1-b}$. Using a round-to-nearest mode, an addition smaller than half that gap will disappear, so that if $e^{-m} < 2^{-b}$, then the floating point representation of $1+e^{-m}$ will be $1$. Thus, under this rounding model, the collapse occurs when $m > b \ln 2$. For bf16, $b=8$, so the threshold is $5.545$; for float32, $b=24$, the threshold is $16.636$; for float64, $b=53$, the threshold is $36.737$. These predicted thresholds line up with the numerical experiments of \citet{barron2026fix}.

Next, let $\vect{z}$ and $\vect{w}$ denote the student and teacher logits, respectively, with student probabilities $\vect{q} = \softmax(\vect{z})$ and target probabilities $\vect{p} = \softmax(\vect{w})$. Holding the teacher logits fixed, in exact arithmetic we have $\nabla_{\vect{z}} \CE(\vect{z}, \vect{p}) = \vect{q} - \vect{p}$. If the student logits approach the teacher logits, we should have that the gradient of the cross-entropy with respect to the student logits approaches 0. So, the second key property we will test is that if the student logits equal the teacher logits, this gradient is also zero. More precisely, for any fixed teacher logits $\vect{w}$,
\begin{align*}
  \left.\nabla_{\vect{z}}\CE(\vect{z},\softmax(\vect{w}))\right|_{\vect{z}=\vect{w}}
  = \vect{0}.
\end{align*}

To see why this can fail, recall that the teacher probabilities are computed using $\softmax(\vect{w})$, while the backward pass of the cross-entropy can compute the student probabilities using $\exp(\logsoftmax(\vect{z}))$. When the logits are identical, these expressions are mathematically equivalent, but their different sequences of floating-point operations can give different rounding errors. 

We can simply test these two properties with Hypothesis, in PyTorch, as below.

\paragraph{Property 1: Each row of the softmax Jacobian sums to zero.}\mbox{}

Here, we compute the Jacobian of the softmax function and check that each row sums to zero with a relative tolerance of $10^{-12}$, using the sum of the absolute entries in that row as the scale. The code is as follows:

\begin{lstlisting}
@given(xs=st.lists(st.floats(allow_nan=False, allow_infinity=False), min_size=2))
@settings(max_examples=10_000, deadline=None)
def test_softmax_gradients_sum_to_zero(softmax, xs):
    x = torch.tensor(xs, dtype=torch.float64, requires_grad=True)
    J = torch.autograd.functional.jacobian(softmax, x)
    assert (J.sum(dim=1).abs() <= 1e-12 * J.abs().sum(dim=1)).all()
\end{lstlisting}

\begin{minipage}{\linewidth}
\paragraph{Property 2: When student and teacher logits match, the cross-entropy gradient with respect to the student logits is zero, holding the teacher fixed.}\mbox{}

Here, \texttt{soft\_target\_cross\_entropy} takes student and teacher logits as its two arguments, and converts the teacher logits to target probabilities internally. Its mathematical loss is therefore $\CE(\vect{z},\softmax(\vect{w}))$. The test passes identical logit values for the student and teacher, with \texttt{x.detach()} keeping the teacher fixed when computing the gradient.

\begin{lstlisting}
@given(xs=st.lists(st.floats(allow_nan=False, allow_infinity=False), min_size=2))
@settings(max_examples=10_000, deadline=None)
def test_soft_target_cross_entropy_has_zero_gradient_for_identical_logits(soft_target_cross_entropy, xs):
    x = torch.tensor(xs, dtype=torch.float64, requires_grad=True)
    loss = soft_target_cross_entropy(x, x.detach())
    grad, = torch.autograd.grad(loss, x)
    assert (grad == 0).all()
\end{lstlisting}
\end{minipage}

These test functions directly test the two properties for a given function (softmax or cross-entropy) and a given tensor. The Hypothesis decorators specify the space of valid inputs, i.e., lists of floats excluding NaN and infinity, with size at least 2, and we generate at most 10,000 such random inputs. The list of floats is then converted to a tensor within the test function. In the implementation in the codebase at \url{https://github.com/mmaaz-git/softmax-fix}, we additionally add a parameter to allow the user to specify which softmax or cross-entropy function to test.

As expected, we find that using the built-in PyTorch implementations of the softmax and cross-entropy functions, these properties fail. On one test run, Hypothesis finds the following failures: the vector $[0.0, 37.0]$ breaks Property 1, and the vector $[0.0, 1.0]$ breaks Property 2. For Property 1, the Jacobian row corresponding to the maximum logit is approximately $[-8.533 \times 10^{-17}, 0]$: the diagonal derivative has collapsed to zero while the other derivative remains nonzero. Its margin of $37$ exceeds the predicted float64 threshold of $36.737$. For Property 2, the computed gradient is approximately $[-5.55 \times 10^{-17}, 0]$, rather than zero.

\section{A Remaining Failure in the Proposed Fix}
\label{sec:remaining-failure}

To fix softmax collapse, \citet{barron2026fix} proposed a fix to $\logsoftmax$ (and compute $\softmax(\cdot)$ as $\exp(\logsoftmax(\cdot))$) and hard-target cross-entropy. They also introduce a new version of the soft-target cross-entropy to fix softmax cancellation. We refer the reader to their paper for the details, which we have also re-implemented in our codebase.

We then test Properties 1 and 2 on their functions. Surprisingly, we find that the test for Property 1 fails, although Property 2 passes. In particular, Hypothesis finds that the following vector produces a NaN Jacobian: $[1.7976931348623157 \cdot 10^{308}, -9.9792015476736 \cdot 10^{291}]$. To see where this failure arises, note that \citep{barron2026fix} do the following to compute $\logsoftmax$: subtract the maximum logit from the logits, explicitly setting the maximum logit to 0; then, only compute the $\logsumexp$ over shifted entries except the maximum logit, and finally apply $\softplus$ to the result; lastly, subtracting this log-normalizer from each shifted logit gives $\logsoftmax$. This fixes the issue that we identified where $1$ plus a small number can round to $1$.

However, in the failure identified by Hypothesis, performing max-shifting overflows the second element to $-\infty$, so when we exclude the maximum we take the $\logsumexp$ of $-\infty$. Although the forward value is correctly $-\infty$, the backward value becomes NaN. Importantly, note that this error does not occur with the built-in PyTorch implementation of softmax. For example, for the vector $[10^{308}, -10^{308}]$, both the standard implementation and the \citet{barron2026fix} implementation correctly give $[1,0]$ for the forward value; but the standard implementation gives all zeros for the backward value, while the \citet{barron2026fix} implementation gives all NaNs.

\section{Our Fix}
\label{sec:our-fix}

We propose a fix for computing $\logsoftmax$ (and hence $\softmax$) which preserves the utility of \citet{barron2026fix}'s fix, but also retains proper gradients in the extreme case we identified above. In our fix, to compute $\logsoftmax$, we perform max-shifting as in \citet{barron2026fix}. But, then, we compute the sum of the exponentials of the shifted non-maximum logits, and then apply $\logonep$ to get the log-normalizer; then, lastly, we subtract this log-normalizer from each shifted logit as usual.

In essence, the difference is the following:

\begin{lstlisting}
# Barron et al.
log_S = logsumexp(others)
normalizer = softplus(log_S)

# Ours
S = exp(others).sum()
normalizer = log1p(S)
\end{lstlisting}

See our codebase for the full implementation. Mathematically, these are equivalent. Let $\vect{a}$ be the shifted logits excluding the selected maximum, and define $S = \sum_i e^{a_i}$. Then, for $S>0$, we have:

\begin{align*}
  \softplus(\logsumexp(\vect{a})) &= \softplus(\log S) \\
  &= \log(1 + e^{\log S}) \\
  &= \log(1 + S) \\
  &= \logonep(S)
\end{align*}

But now note that for $S=0$, the derivative of $\logonep(S)$ is finite and equals $1$, whereas automatic differentiation of $\logsumexp$ on an all-$-\infty$ vector gives NaN. Indeed, our implementation actually does pass the tests for Properties 1 and 2. To see this concretely, consider the vector $[10^{308}, -10^{308}]$. Our implementation gives the forward value $[1,0]$, and the backward value all zeros, as seen in Table \ref{tab:overflow-comparison}.

\begin{table}[H]
  \caption{Forward softmax probabilities and computed Jacobians for
  $[10^{308}, -10^{308}]$, evaluated in float64 on CPU using PyTorch 2.7.1.}
  \label{tab:overflow-comparison}
  \centering
  \begin{tabular}{lcc}
    \toprule
    Implementation & Forward probabilities & Computed Jacobian \\
    \midrule
    Standard (PyTorch) & $[1,0]$ & All zeros \\
    \citet{barron2026fix} (PyTorch translation) & $[1,0]$ & All NaNs \\
    Ours & $[1,0]$ & All zeros \\
    \bottomrule
  \end{tabular}
\end{table}

We reuse \citet{barron2026fix}'s fix for soft-target cross-entropy, only changing the $\logsoftmax$ computation. We also note that using $\logonep$ is not necessarily a new idea: \citet{blanchard2021accurately} also use it to compute $\logsumexp$. However, our implementation combines this with \citet{barron2026fix}'s explicit handling of the maximum logit.

The root of the \citet{barron2026fix} failure is taking the $\logsumexp$ of an all-$-\infty$ vector. While this can certainly occur with enormous finite-logit margins, this also appears practically in e.g., causal attention masking in transformers \citep{vaswani2017attention}. Precisely, the first query can attend only to the first key, giving a masked logit row that looks like $[a, -\infty, -\infty, \ldots, -\infty]$, where $a$ is finite. Excluding the maximum logit gives an all-$-\infty$ vector, who's $\logsumexp$ is $-\infty$, but whose backward gives NaN gradients. So, using \citet{barron2026fix}'s softmax for the attention probabilities in this case would produce NaN gradients, whereas our fix gives the correct forward and backward values. 

\section{Conclusion}
\label{sec:conclusion}

We reframed the numerical issues found by \citet{barron2026fix} as failures of simple mathematical properties of the softmax and cross-entropy functions, and wrote property-based tests that rediscover these issues. However, we then find that their proposed softmax fails the Jacobian row-sum test by producing NaNs for some finite inputs. We then developed a new fix, using $\logonep$, which makes it more robust to overflow after max-shifting, and also adequately handles the case of causal attention masking. More generally, we hope that property-based testing can be adopted more widely in test suites in machine learning libraries to discover such numerical issues.

\newpage
\bibliographystyle{plainnat}
\bibliography{references}

\end{document}

%% file: preamble.tex
\usepackage[T1]{fontenc}
\usepackage[utf8]{inputenc}
\usepackage{microtype}
\usepackage{amsmath,amssymb,amsthm}
\usepackage{graphicx}
\usepackage{flafter}
\usepackage{booktabs}
\usepackage{algorithm}
\usepackage{algpseudocode}
\usepackage{listings}
\usepackage{xcolor}
\definecolor{codebackground}{gray}{0.97}
\definecolor{codeborder}{gray}{0.82}
\definecolor{codecomment}{gray}{0.50}
\usepackage[hidelinks]{hyperref}
\usepackage[capitalize,noabbrev]{cleveref}

\graphicspath{{figures/}}

\definecolor{placeholdergray}{gray}{0.38}

\newcommand{\R}{\mathbb{R}}

\newcommand{\vect}[1]{\boldsymbol{#1}}

\DeclareMathOperator{\softmax}{softmax}
\DeclareMathOperator{\logsoftmax}{logsoftmax}
\DeclareMathOperator{\logsumexp}{logsumexp}
\DeclareMathOperator{\logonep}{log1p}
\DeclareMathOperator{\softplus}{softplus}
\DeclareMathOperator{\CE}{CE}

\theoremstyle{definition}

\theoremstyle{remark}

%% file: references.bib
@inproceedings{vaswani2017attention,
  author    = {Vaswani, Ashish and Shazeer, Noam and Parmar, Niki and Uszkoreit, Jakob and Jones, Llion and Gomez, Aidan N. and Kaiser, {\L}ukasz and Polosukhin, Illia},
  title     = {Attention Is All You Need},
  booktitle = {Advances in Neural Information Processing Systems},
  volume    = {30},
  year      = {2017},
  url       = {https://papers.nips.cc/paper_files/paper/2017/hash/3f5ee243547dee91fbd053c1c4a845aa-Abstract.html}
}

@book{higham2002accuracy,
  author    = {Nicholas J. Higham},
  title     = {Accuracy and Stability of Numerical Algorithms},
  publisher = {Society for Industrial and Applied Mathematics},
  edition   = {2},
  year      = {2002}
}

@article{blanchard2021accurately,
  author    = {Blanchard, Pierre and Higham, Desmond J. and Higham, Nicholas J.},
  title     = {Accurately computing the log-sum-exp and softmax functions},
  journal   = {IMA Journal of Numerical Analysis},
  volume    = {41},
  number    = {4},
  pages     = {2311--2330},
  year      = {2021},
  doi       = {10.1093/imanum/draa038},
  url       = {https://doi.org/10.1093/imanum/draa038}
}

@misc{prieto2025grokking,
  title         = {Grokking at the Edge of Numerical Stability},
  author        = {Prieto, Lucas and Barsbey, Melih and Mediano, Pedro A. M. and Birdal, Tolga},
  year          = {2025},
  eprint        = {2501.04697},
  archivePrefix = {arXiv},
  primaryClass  = {cs.LG},
  doi           = {10.48550/arXiv.2501.04697},
  url           = {https://arxiv.org/abs/2501.04697}
}

@misc{barron2026fix,
  title         = {A Fix for Softmax and Cross-Entropy},
  author        = {Barron, Jonathan T. and Gnanapragasam, Dan and Tschannen, Michael},
  year          = {2026},
  eprint        = {2610.31642},
  archivePrefix = {arXiv},
  primaryClass  = {cs.LG},
  doi           = {10.48550/arXiv.2610.31642},
  url           = {https://arxiv.org/abs/2610.31642}
}

@misc{barron2026softmax_tweet,
  author       = {Barron, Jonathan T.},
  title        = {Tweet about softmax cancellation and softmax collapse},
  year         = {2026},
  month        = oct,
  howpublished = {X (formerly Twitter)},
  url          = {https://x.com/jon_barron/status/2113707885486375022},
  note         = {Posted October 23, 2026. Accessed October 25, 2026}
}

@inproceedings{claessen2000quickcheck,
  author    = {Claessen, Koen and Hughes, John},
  title     = {{QuickCheck}: A Lightweight Tool for Random Testing of {Haskell} Programs},
  booktitle = {Proceedings of the Fifth ACM SIGPLAN International Conference on Functional Programming},
  series    = {ICFP '00},
  pages     = {268--279},
  publisher = {Association for Computing Machinery},
  year      = {2000},
  doi       = {10.1145/351240.351266},
  url       = {https://doi.org/10.1145/351240.351266}
}

@article{maciver2019hypothesis,
  author  = {MacIver, David R. and Hatfield-Dodds, Zac and {many other contributors}},
  title   = {{Hypothesis}: A New Approach to Property-Based Testing},
  journal = {Journal of Open Source Software},
  volume  = {4},
  number  = {43},
  pages   = {1891},
  year    = {2019},
  doi     = {10.21105/joss.01891},
  url     = {https://joss.theoj.org/papers/10.21105/joss.01891}
}

@misc{maaz2025agentic,
  author        = {Maaz, Muhammad and DeVoe, Liam and Hatfield-Dodds, Zac and Carlini, Nicholas},
  title         = {Agentic Property-Based Testing: Finding Bugs Across the {Python} Ecosystem},
  year          = {2025},
  eprint        = {2510.09907},
  archivePrefix = {arXiv},
  primaryClass  = {cs.SE},
  doi           = {10.48550/arXiv.2510.09907},
  url           = {https://arxiv.org/abs/2510.09907},
  note          = {Presented at the 4th Deep Learning for Code Workshop, NeurIPS 2025}
}
